\documentclass[letterpaper]{article} 
\usepackage{aaai25}
\usepackage{times}  
\usepackage{helvet}  
\usepackage{courier}  
\usepackage[hyphens]{url}  
\usepackage{graphicx} 
\usepackage{natbib}  
\usepackage{caption} 
\usepackage{algorithm}
\usepackage{algorithmic}
\usepackage{multirow} 
\usepackage{calc}
\usepackage{caption}

\usepackage{newfloat}
\usepackage{listings}
\DeclareCaptionStyle{ruled}{labelfont=normalfont,labelsep=colon,strut=off} 
\floatstyle{ruled}
\newfloat{listing}{tb}{lst}{}
\floatname{listing}{Listing}
\usepackage{latexsym}
\usepackage{amssymb}
\usepackage{amsmath}
\usepackage{amsthm}
\usepackage{booktabs}
\usepackage{enumitem}
\usepackage{graphicx}
\usepackage{color}

\usepackage{eqparbox}
\usepackage{xspace}
\usepackage[usenames,dvipsnames,svgnames]{xcolor}

\newcommand{\mr}[2]{\multirow{#1}{*}{#2}}

\newcommand{\red}[1]{{\textcolor{red}{#1}}}

\newcommand{\citeme}[1]{\red{[XX]}}
\newcommand{\refme}[1]{\red{(XX)}}

\newcommand{\real}{\mathbb{R}}

\newcommand{\defn}{\mathrel{:=}}

\newcommand{\cD}{\mathcal{D}}

\newcommand{\cL}{\mathcal{L}}

\newcommand{\vB}{\mathbf{B}}

\newcommand{\vD}{\mathbf{D}}

\newcommand{\vM}{\mathbf{M}}

\newcommand{\vP}{\mathbf{P}}

\newcommand{\vb}{\mathbf{b}}
\newcommand{\vc}{\mathbf{c}}

\newcommand{\vg}{\mathbf{g}}

\newcommand{\vt}{\mathbf{t}}

\newcommand{\vv}{\mathbf{v}}

\newcommand{\vy}{\mathbf{y}}

\makeatletter
\newcommand*\bdot{\mathpalette\bdot@{.7}}
\newcommand*\bdot@[2]{\mathbin{\vcenter{\hbox{\scalebox{#2}{$\m@th#1\bullet$}}}}}
\makeatother

\makeatletter
\DeclareRobustCommand\onedot{\futurelet\@let@token\@onedot}
\def\@onedot{\ifx\@let@token.\else.\null\fi\xspace}

\def\ie{\emph{i.e}\onedot} 
 
 \def\vs{\emph{vs}\onedot}
\def\wrt{w.r.t\onedot}  

\makeatother

\usepackage{tikz}
\usetikzlibrary{arrows,shapes,calc,matrix,fit,backgrounds}
\usepackage{pgfplots}
\usepackage{pgfplotstable}
\pgfplotsset{compat=1.9}

\usepackage{xstring}

\usepgfplotslibrary{external}

\IfBeginWith*{\jobname}{fig/extern/}{\finalcopy}{}

\newcommand{\leg}[1]{\addlegendentry{#1}}

\tikzset{every mark/.append style={solid}}
\pgfplotsset{
	grid=both, width=\columnwidth, try min ticks=5,
	every axis/.append style={font=\scriptsize},
	every axis plot/.append style={thick,mark=none,mark size=1.2,tension=0.18},
	legend cell align=left, legend style={fill opacity=0.8},
}

\pgfplotsset{
	dash/.style={mark=o,dashed,opacity=0.7},
	dott/.style={mark=o,dotted,opacity=0.7},
}

\title{Revisiting Physically Realizable Adversarial Object Attack against LiDAR-based Detection: Clarifying Problem Formulation and Experimental Protocols}

\author{
    Luo Cheng \textsuperscript{\rm 1}\textsuperscript{\rm 2}, Hanwei Zhang\textsuperscript{\rm 3}\textsuperscript{\rm 4}, Lijun Zhang\textsuperscript{\rm 2}, Holger Hermanns\textsuperscript{\rm 3}
}
\affiliations{
    \textsuperscript{\rm 1}University of Chinese Academy of Sciences, Beijing, China\\
    \textsuperscript{\rm 2}Key Laboratory of System Software (Chinese Academy of Sciences) and State Key Laboratory of Computer Science, Institute of Software, Chinese Academy of Sciences, Beijing, China\\
    \textsuperscript{\rm 3}Universität des Saarlandes, Saarland, Germany\\
    \textsuperscript{\rm 4}Institute of Intelligent Software, Guangzhou, China\\

}

\begin{document}

\maketitle

\begin{abstract}
Adversarial robustness in LiDAR-based 3D object detection is a critical research area due to its widespread application in real-world scenarios. While many digital attacks manipulate point clouds or meshes, they often lack physical realizability, limiting their practical impact. Physical adversarial object attacks remain underexplored and suffer from poor reproducibility due to inconsistent setups and hardware differences. To address this, we propose a device-agnostic, standardized framework that abstracts key elements of physical adversarial object attacks, supports diverse methods, and provides open-source code with benchmarking protocols in simulation and real-world settings. Our framework enables fair comparison, accelerates research, and is validated by successfully transferring simulated attacks to a physical LiDAR system. Beyond the framework, we offer insights into factors influencing attack success and advance understanding of adversarial robustness in real-world LiDAR perception.
\end{abstract}

\section{Introduction}

Adversarial phenomena \cite{kurakin2018adversarial, zhang2022deep} pose a significant threat to deep learning models. In particular, ensuring adversarial robustness in LiDAR-based 3D object detection is crucial due to its widespread use in real-world applications across various industries, including autonomous vehicles, delivery robots, warehouse automation, and drones~\cite{aung2024review}. 
While adversarial attacks against 3D point cloud models by inserting, deleting, or shifting points \cite{si_adv_pc, geoa3, zhang2024eidos} or modifying the reconstructed meshes \cite{mesh-attack} from 3D point cloud, have advanced our understanding of model vulnerabilities, many of these methods are not physically realizable. In practice, such manipulations are often impractical or easily overlooked. While idealized studies of adversarial robustness help identify model flaws, evaluating physical attacks is crucial for assessing real-world robustness and ensuring AI systems meet safety standards like those in the EU AI Act\footnote{Regulation (EU) 2024/1689 on Artificial Intelligence (AI Act), \url{https://eur-lex.europa.eu/legal-content/EN/TXT/?uri=OJ:L_202401689}}.

Despite growing interest, physical attacks on LiDAR-based 3D object detection remain underexplored compared to digital attacks \cite{zhang2024comprehensive}. Existing physical attacks fall into two types: spoofing attacks, which inject fake points into LiDAR streams via lasers or reflectors \cite{cao2019adversarial, sun2020towards}, and adversarial objects, \ie crafted physical artifacts to mislead perception systems \cite{tu2020physically, abdelfattah2021towards}. We view adversarial object attacks as a serious threat, but research is limited due to poor reproducibility. Few studies share code or setups, and hardware differences hinder fair comparison. To address this, we propose a device-agnostic, standardized framework to enable reproducible benchmarking and accelerate research progress.

\begin{figure}[t]
\centering
\setlength{\tabcolsep}{1pt}
\begin{tabular}{c c}

 \includegraphics[width=0.45\linewidth, height=0.35\linewidth]{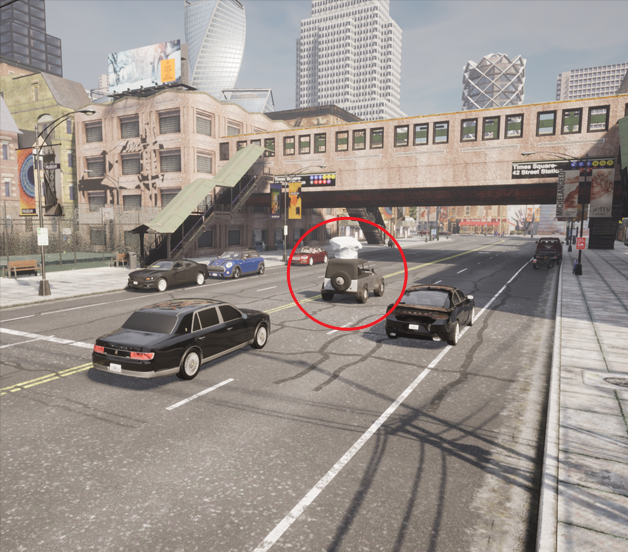}
& \includegraphics[trim={12mm 22mm 12mm 20mm},clip,width=0.45\linewidth, height=0.35\linewidth]{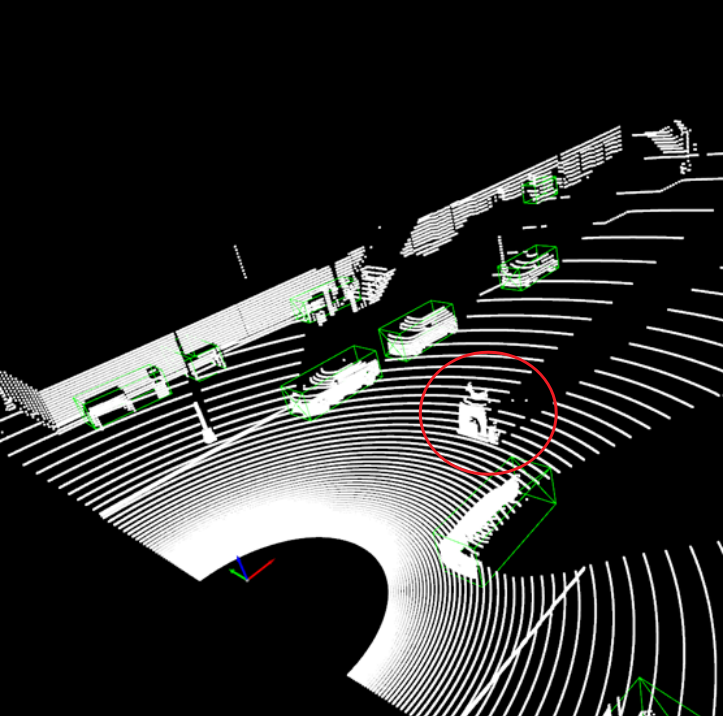}    \\ 
\includegraphics[trim=400 50 400 50, clip, width=0.45\linewidth, height=0.35\linewidth]{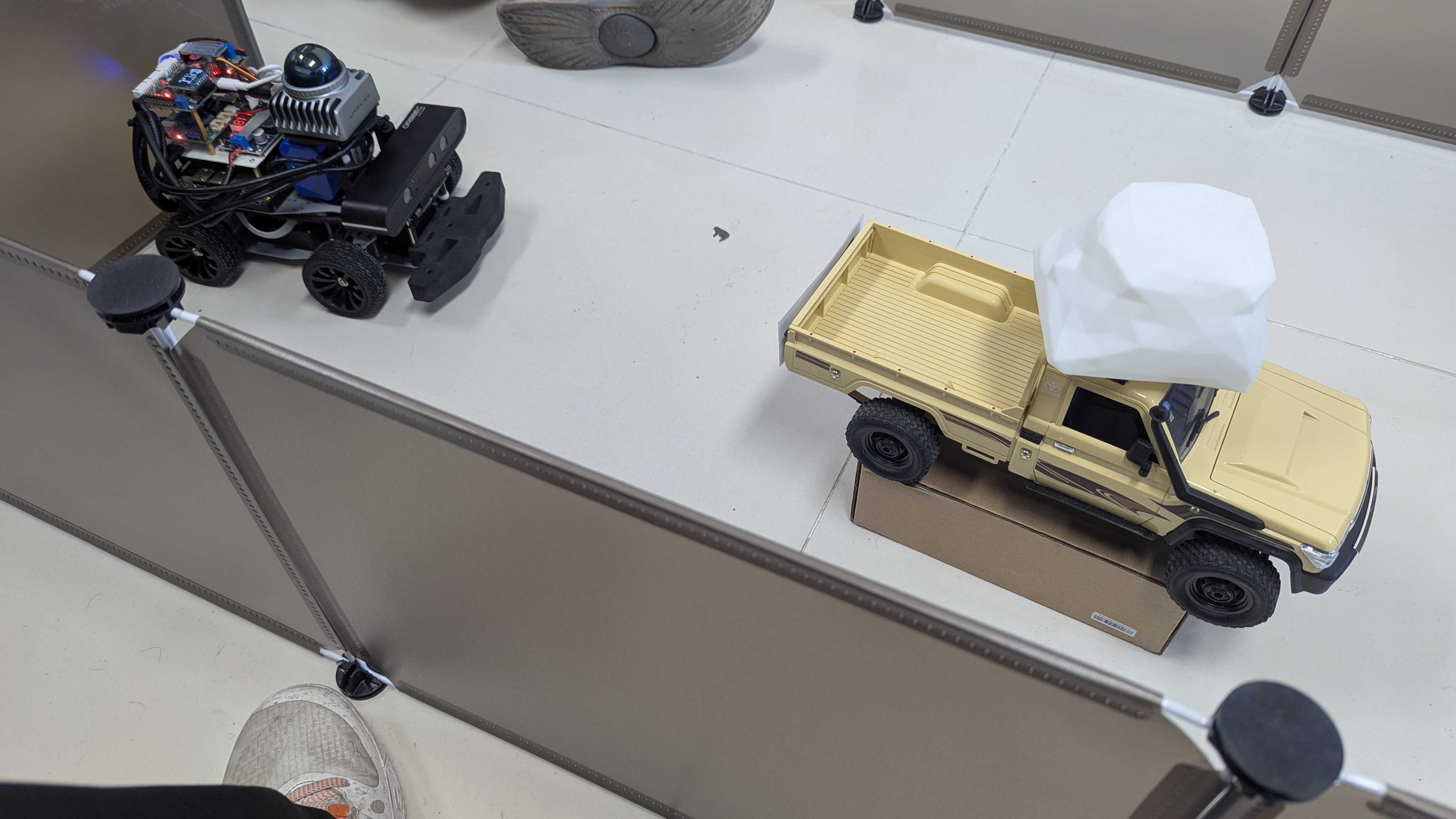}
&
\includegraphics[trim=80 50 40 50, clip, scale=0.7, width=0.45\linewidth, height=0.35\linewidth]{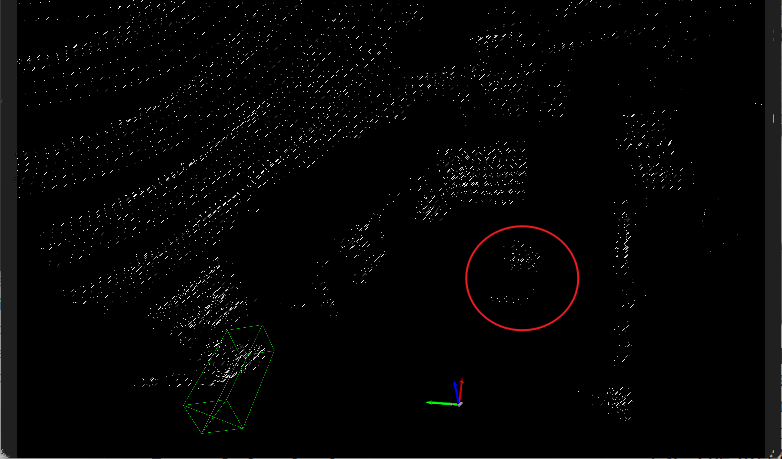}
\end{tabular}
\caption{Top row shows the simulated scenario in CarLA; bottom row shows the real-world setup with a small car and a 3D-printed adversarial object. Left column presents the camera view of the scene, including the car with the adversarial object and the LiDAR-equipped car. Right column shows the corresponding LiDAR point cloud perception by the LiDAR-equipped car.}

\label{fig:intro-fig}
\end{figure}

Our standardized, device-agnostic framework abstracts the key components of physically realizable adversarial object attacks. By adjusting elements like mesh shape, loss constraints, and object placement, our framework can encompass existing attack methods. To enable fair comparison, we provide open-source code with evaluation protocols in both simulation and dataset settings. Figure~\ref{fig:intro-fig} validates that our simulation setting closely approximates the real world. We 3D-print a simulated adversarial object and demonstrate its real-world effectiveness on a physical LiDAR-based detection system. 
Our framework supports various loss functions, optimization strategies, and attack scenarios (white/black-box), and is compatible with multiple detection models. In this work, we follow a widely adopted setup, using it as a case study to analyze key factors affecting attack success and to provide foundational insights into physical adversarial objects. Thus, our contributions go beyond introducing a framework with benchmarkable settings, evaluation protocols, and open-source code to advance research on physical adversarial objects. We also provide insights and preliminary knowledge on real-world LiDAR robustness.


\section{Related Work}

\subsection{Robustness of LiDAR-based 3D Object Detection}

The robustness of LiDAR-based 3D object detection typically includes stability under varying conditions~\cite{song2024robustness} and adversarial robustness, \ie, resilience against deliberate attacks~\cite{zhang2024comprehensive}, with recent focus on the latter. 
\citet{zhang2024comprehensive} benchmark detectors by perturbing 3D points and categorize them as voxel-based~\cite{yan2018second, lang2019pointpillars, deng2021voxel}, point-based~\cite{shi2019pointrcnn}, or mixed-feature~\cite{shi2020pv}, as well as one-stage~\cite{yan2018second, lang2019pointpillars} or two-stage~\cite{shi2020pv,shi2019pointrcnn} architectures.
We follow this categorization but evaluate robustness against physically realizable attacks by introducing adversarial objects into the scene, better reflecting real-world threats in critical applications like autonomous driving.

\subsection{Adversarial Attacks in Physical World}

Physical-world adversarial attacks have received increasing attention due to their real-world safety implications \cite{ren2021adversarial}, spanning image classification \cite{li2019adversarial}, pedestrian detection \cite{thys2019fooling}, infrared imaging \cite{zhou2018invisible}, and LiDAR-based perception \cite{tu2020physically, cao2019adversarial}. Among these, attacks on LiDAR systems are particularly important given their widespread use in autonomous driving. 

Attacks on LiDAR systems can be broadly categorized into two types: spoofing attacks and adversarial objects. Spoofing attacks~\cite{cao2019adversarialSpoof} target the LiDAR sensor by injecting fake points to hinder detection, simulate fake vehicles, or disrupt vehicle networks and communication~\cite{ju2022survey}. However, their effectiveness is often limited in real-world scenarios due to environmental factors like weather, sensor noise, and difficulties in reproducing results outside controlled conditions. We consider adversarial objects to pose a greater threat in real-world applications.


\paragraph{Adversarial Object.}
PhyAdv \cite{tu2020physically} showed that placing an adversarial object on a vehicle's rooftop can effectively hide it from LiDAR detectors. Similarly, \citet{abdelfattah2021towards} extended adversarial techniques from 2D to 3D detection by using RGB differentiable rendering for adversarial object generation. Inspired by object placement, \cite{zhu2021can} proposed an attack framework that places reflective objects in physical space to mislead LiDAR sensing systems.  \citet{wang2023adversarial} developed an adversarial obstacle generation algorithm to deceive deep 3D detection models by creating fake obstacles within LiDAR’s detection range using only a few points. 
These studies demonstrate the feasibility and severity of adversarial object threats in realistic settings. However, inconsistent techniques, varied setups, and limited model coverage hinder the development of general insights into adversarial robustness in LiDAR-based perception. To address this, we build on prior work with a generalizable framework and comprehensive evaluation.

\section{Physically Realizable Adversarial Object}

\paragraph{Problem Formulation: Adversarial Object.} 

Let $\vP$ denote an unordered set of 3D points and $f$ denote a LiDAR-based 3D object detection which maps a point cloud $\vP$ to a set of 3D object bounding boxes $\vB$ associated with the corresponding probability scores $\vs$ of the predicted class labels $\vy$.
Let $\vM \defn \{\vv_i\}$ denote a mesh object constructed by the vertices $\vv_i$.
Let $G(\vP, \vM, \vt)$ denote a function that integrates a mesh $\vM$ into a point cloud $\vP$ at given position and orientation $\vt$, producing the resulting point clouds $\vP'$. Möller-Trumbore intersection algorithm~\cite{moller2005fast} is a solution for $G$, which computes ray-triangle intersections between LiDAR rays and the mesh faces.


We aim to find an adversarial mesh $\vM_{adv}$ against the target LiDAR-based 3D object detection $f$ over the 3D points $\vP$ from the distribution $\cD$, satisfying
\begin{align}
\footnotesize
    \arg\min_{\vM_{adv}} &~~ \sum_{\vP \sim \cD} \cL(f, G(\vP, \vM_{adv}, \vt)), \label{eq:mis_adv} \\
    s.t. &~~ \Phi(\vM_{adv}) = True \label{eq:phy_fea}
\end{align}
where $\cL(f, G(\vP, \vM_{adv}, \vt))$ is the misdetection loss and $\Phi(\vM)$ is a boolean function that determines if a mesh $\vM$ is physically feasible. 

\paragraph{Positioning of Object.}
The placement of the object, \ie $\vt$, depends on the overall design and physical context. Several studies place the adversarial object on top of the car, as this position ensures greater visibility across diverse scenarios and poses a more significant threat.

\paragraph{Physically Feasible Constraint.}
Implementing directly the boolean function $\Phi$ to check the physical feasibility of a mesh is challenging. To enforce this constraint during optimization, two main strategies are used: (1) reparameterizing the adversarial mesh to implicitly include the constraint; (2) incorporating a measure of physical feasibility into the loss function.

An example of parameterization is from PhyAdv~\cite{tu2020physically}, which reformulates the problem using learnable local displacement vectors $\Delta \vv_i \in \real^3$ for each vertex. 
To enforce a box constraint and ensure physical feasibility with given size and translation limits, we denote the reparameterization as
\begin{equation}
\vv_i = \tau(\vv_i^0, \Delta \vv_i),
    \label{eq:box-constraint}
\end{equation}
where $\vv_i^0$ is the initial vertex.
On the other hand, surface smoothness can serve as a proxy for measuring physical feasibility of a mesh. Accordingly, the Laplacian loss for estimating surface smoothness
is defined as: 
\begin{equation}
\footnotesize
    \phi(\vM_{adv}) \defn \sum_{\vv_i \in V} \| \vv_i - \frac{1}{\|N(i)\|} \sum_{j \in N(i)} \vv_j \|_2^2,
    \label{eq:laplacian-loss}
\end{equation} 
where $V$ is the set of vertices in $\vM_{adv}$ and $N(i)$ denotes the neighboring vertices of $\vv_i$.



\paragraph{Adversarial Misdetection Loss.}
The misdetection loss (\ref{eq:mis_adv}) is critical to the effectiveness of adversarial objects and should be tailored to the threat objective and model architecture. Existing studies on physical adversarial objects often overlook this aspect, typically defaulting to loss functions tied to the initial stage of two-stage detectors. 
Since detection involves both localization and classification, their individual contributions to adversarial effectiveness remain underexplored. To address this, we examine different loss formulations and group them into three categories: mislocalization (ML) loss, misrecognition (MR) loss, and comprehensive misdetection (C) loss, aiming to enhance attack performance across architectures.


\emph{Mislocalization Loss (ML):} Misdetection occurs when the detection model fails to correctly locate the bounding boxes, regardless of the predicted class. The misdetection loss can be formulated as
\begin{equation}
\footnotesize
    \cL(f,\cD,\vM_{adv}) \defn \sum_{\vP \sim \cD} \sum_k IoU \left(\vB,B_{gt}\right),
    \label{eq:iou-loss}
\end{equation}
where $\vB$ is the proposed bounding boxes and $B_{gt}$ denotes the given ground truth. A proposal is considered relevant if both its confidence score and its IoU with ground truth bounding box are greater than $0.1$. This can be further extended by weighting each bounding box using its fixed prediction score.

\emph{Misrecognization Loss (MR):} Misrecognition is also considered misdetection, meaning the object is correctly localized but not correctly recognized. A straightforward definition is
\begin{equation}
\footnotesize
\cL(f,\cD,\vM_{adv}) \defn \sum_{\vP \sim \cD} \sum_k -\log\left(1-\sigma(s))\right).
    \label{eq:logit-loss}
\end{equation}
where $s$ is the prediction logit of the target class, $\sigma(\cdot)$ is the sigmoid function and $\sigma(s)$ is prediction score from the network. A weighted version is used in PhyAdv~\cite{tu2020physically}:
\begin{equation}
\footnotesize
   \cL(f,\cD,\vM_{adv}) \defn \sum_{\vP \sim \cD} \sum_k -IoU \left(\vB,B_{gt}\right) \log\left(1-\sigma(s))\right),
    \label{eq:full-loss}
\end{equation} 
where the predicted bounding box and its corresponding IoU value are kept fixed as weights for the confidence scores.

\emph{Comprehensive Misdetection Loss (C):} To encompass both mislocalization and misrecognition in misdetection, we use (\ref{eq:full-loss}) without fixing either the proposed bounding boxes or the prediction scores. However, it is ineffective in the second stage of a two-stage model when the prediction score is nearly one. Thus, for such a case, we propose to construct the loss as 

{\footnotesize
\begin{align}
    \cL(f,\cD,\vM_{adv}) \defn &\sum_{\vP \sim \cD} \sum_k IoU \left(\vB,B_{gt}\right) \cdot \sigma(s).
    \label{eq:full-loss1}
\end{align}
}

\emph{Logit \vs Score.} Logit loss is simple yet effective for transferable adversarial attacks in images~\cite{zhao2021success}. However, in 3D detection attacks, only the score is used as part of the loss. To fill the gap, we modify (\ref{eq:full-loss1}) as

{
\footnotesize
\begin{align}
    \cL(f,\cD,\vM_{adv}) \defn & \sum_{\vP \sim \cD} \sum_k IoU \left(\vB,B_{gt}\right) \cdot s,
    \label{eq:full-loss-logit}
\end{align}
}and compare the performance between using the score and the logit of object prediction.


\paragraph{A Study Case.}
Our framework is designed to flexibly support different initialization shapes and placements of adversarial objects, enabling it to encompass a wide range of existing work. While variations in shape, position, and constraints can significantly alter the physical representation of an attack, we emphasize that the core challenge lies in the algorithmic components—specifically, how to formulate effective physically feasible constraints, define misdetection losses, and solve the optimization problem efficiently. These components remain relevant regardless of the object’s initial shape or placement. Therefore, for clarity and consistency, we adopt a standard setup: a mesh initialized as a sphere, positioned on top of a car.

To generate a physically realizable adversarial object $\vM_{adv}$, we formulate an optimization problem,  
\begin{equation}
\footnotesize
    \ell(f,\cD,\vM_{adv}) \defn \cL(f,\cD,\vM_{adv}) + \lambda \phi(\vM_{adv}),
    \label{eq:loss_all}
\end{equation}
that balances the misdetection loss and the physical feasibility constraints by a weighting parameter $\lambda$.

We generalize the attack algorithm as shown in Algorithm \ref{alg:lidarAdv}. We use Gradient Descent (GD) with fixed step size $\epsilon$ as the optimizer. The simple yet efficient framework enables attacks in both white-box and black-box settings and allows the exploration of various definitions of misdetection loss $\cL$.

\begin{algorithm}[htbp]
\small
\caption{ \small{Adversarial Attack against LiDAR-based Detector}}
\label{alg:lidarAdv}
\textbf{Input:} $g$: Targeted LiDAR-based 3D Detector, $\vD$: Given dataset \\
\textbf{Input:} $f$: (Optional) Surrogate LiDAR 3D Detector \\
\textbf{Input:} $\nu$: Sphere level, $\vb$: Scale, $\epsilon$: Step size \\
\textbf{Output:} $\vM_{adv} \defn  \{\tau(\vv_i^0, \Delta {\vv}_i)\}:$ An adversarial object defined by attacked vertices.
\begin{algorithmic}[1]
        \STATE ${\vv_i^0} \gets Sphere(\nu, \vb)$, $ \Delta {\vv}_i \gets \mathbf{0}$
        \STATE $\vM_{adv} \gets \{\tau(\vv_i^0, \Delta {\vv}_i)\}$ \wrt (\ref{eq:box-constraint})
	\WHILE{$\vP \in \vD$}	       
        \IF{the attack is under white-box setting}
            \STATE $\vg \gets \nabla_{\Delta {\vv}_i}\ell(g,\vP,\vM_{adv})$ \wrt (\ref{eq:loss_all})
            \STATE $\Delta {\vv}_i \gets \Delta {\vv}_i - \epsilon \frac{\vg}{\|\vg \|_2}$
            \STATE $\vM_{adv} \gets \{\tau(\vv_i^0, \Delta {\vv}_i)\}$ \wrt (\ref{eq:box-constraint})
        \ELSIF{the attack is under black-box setting}
            \STATE $\vg \gets \nabla_{\Delta {\vv}_i}\ell(f,\vP,\vM_{adv})$ \wrt (\ref{eq:loss_all})
            \IF{$\cL(g,\vP, \{\tau(\vv_i^0, \Delta \vv_i)\}) < \cL(g, \vP, \vM_{adv}) $ }
            \STATE $\Delta \vv_i  \gets \Delta {\vv}_i - \epsilon \frac{\vg}{\|\vg \|_2}$
            \STATE $\vM_{adv} \gets \{\tau(\vv_i^0, \Delta {\vv}_i)\}$ \wrt (\ref{eq:box-constraint})
            \ENDIF
        \ENDIF
	\ENDWHILE     
\end{algorithmic}
\end{algorithm}

In line 1, we initialize the mesh as a isotropic sphere with parameters sphere level $\nu$ and scale $\vb$. The initial learnable displacement vector $\Delta {\vv}_i$ is set to zero. In line 5 and 9, we estimate the adversarial loss of LiDAR-based 3D detection $f$ on 3D points $\vP$ with adversarial mesh $\vM_{adv}$, then calculate the gradient with respect to the learnable vector $\Delta {\vv}_i$. In the black-box setting, $f$ is the surrogate 3D detector, whereas in the white-box setting, $g$ serves as the sole target detector.
Gradient descent is applied on line 6 and 11. In a white-box setting, we update the adversarial mesh at each step. 
However, in a black-box setting, we update the adversarial mesh only when the misdetection loss on the targeted detection model $g$ decreases.

\section{Evaluation Protocol}

Attack success rate and invisibility are crucial in attacks. For LiDAR-based 3D object detection, precision is measured via 3D bounding boxes or BEV, though not all evaluations consider both. Assessing adversarial object invisibility is also challenging. We propose a standardized evaluation protocol to thoroughly assess adversarial attack performance and detector robustness.

\paragraph{Attack Success Rate.}
We consider two precisions: 3D mean average precision (3D mAP) and Bird's Eye View mean average precision (BEV mAP), which projects 3D bounding boxes onto a 2D plane. Both are measured with moderate difficulty with an IoU threshold of $0.7$. To calculate the attack success rate (ASR), we subtract the accuracy with the adversarial mesh ($p_a$) from the original precisions ($p_o$) and divide by the original precisions, \ie $ASR \defn \frac{p_o - p_a}{p_o}$. 

\paragraph{Invisibility.}
When evaluating the invisibility of an adversarial object, we consider three key aspects: i) \emph{modification magnitude}, measured by the $L_2$ norm of changes to mesh vertices comparing to initialization; ii) \emph{smoothness}, measured by the Laplacian smoothness ($\phi$) of the vertices, reflecting physical realism; iii) \emph{size of object}, measure by two metrics, \ie \emph{Area} and \emph{Volume}. \emph{Area} is the BEV area of the adversarial mesh, calculated as the union of triangle areas projected onto the BEV view, formalized as
\begin{equation}
\footnotesize
\text{Area}(\vM_{adv}) \defn \bigcup_{ \{\vv_i,\vv_j,\vv_k\} \subset \vM_{adv}} \text{Proj}_\text{xy}\ (\{\vv_i,\vv_j,\vv_k\}),
    \label{eq:projected-area}
\end{equation} 
where $\text{Proj}_\text{xy}$ calculates the area of a triangle projected onto the BEV view using only the $x$ and $y$ axes. \emph{Volume} is the total volume of the mesh, computed using the signed volume of tetrahedron algorithm, formalized as
\begin{equation}
\footnotesize
\text{Volume}(\vM_{adv}) \defn | \sum_{ \begin{subarray}{l}\{\vv_i,\vv_j,\vv_k\} \text{ is} \\  \text{a face in}~\vM_{adv}\end{subarray}} \frac{1}{6} (\vv_i \cdot (\vv_j \times \vv_k  ))|,
    \label{eq:signed-volume}
\end{equation} 
\ie taking the absolute value of the sum of the signed volumes for all mesh tetrahedra defined by triangular faces. The first two aspects are standard measures of invisibility, while size is specifically tailored to adversarial objects.

\paragraph{Physical Realizable Setting.}
We structure our experimental setup across three levels: dataset-based evaluation, simulation, and limited real-world testing. The core of our experimental protocol focuses on the dataset and simulation levels, which provide controlled, reproducible, and scalable environments for benchmarking.
To assess real-world applicability, we conduct a small-scale physical experiment demonstrating the transferability of adversarial objects from simulation to reality, as shown in Figure~\ref{fig:intro-fig}. However, due to the complexity, cost, and lack of standardization in physical setups, we do not include real-world testing as part of our formal evaluation protocol.

Our standard procedure follows a train-test split: adversarial objects are generated using the training set and evaluated on the test set. While this structure is widely adopted, we emphasize that assessing physical realizability requires conditions that closely approximate the real world. Since conducting physical tests with actual vehicles is logistically demanding and difficult to reproduce, we adopt a simulation-based approach as a practical surrogate. In our setting, the attacker generates adversarial objects using a dataset aligned in format with the simulation data but drawn from different samples. These objects are then evaluated within a simulation environment, allowing for consistent, scalable, and physically meaningful testing without the constraints of real-world deployment.


\section{Experimental Settings}

\paragraph{Dataset.}
We generate and test adversarial meshes using the KITTI~\cite{geiger2012we} dataset, which provides 3D point clouds from a 64-line Velodyne HDL-64E LiDAR and object labels from the front camera of autonomous vehicles. Focusing on the ``Car" class, we consider objects with over ten points per scene. Out of 28,742 labeled cars in our training and validation datasets, 25,844 meet this criterion.
We evaluate the transferability of adversarial meshes using the KITTI-CARLA (CarLA)~\cite{deschaud2021kitticarla} dataset. Since it only provides semantic labels, we first use the victim model to generate 3D bounding boxes from clean data. Ground truth boxes are those that overlap with ``Car" points by more than $70\%$.

.


\paragraph{Networks.}
We select PointRCNN (PR)~\cite{shi2019pointrcnn}, PointPillar (PP)~\cite{lang2019pointpillars}, PV-RCNN (PVR)~\cite{shi2020pv}, Voxel RCNN (VR)~\cite{deng2021voxel}, and SECOND (SC)~\cite{yan2018second} as representative models for LiDAR-based 3D object detectors\footnote{Codes and weights pre-trained on KITTI are from OpenPCDet\cite{openpcdet2020}. (\url{https://github.com/open-mmlab/OpenPCDet}) }.
The original BEV mAP on KITTI is $86.76\%$ for PP, $85.67\%$ for PR, $88.92\%$ for PVR, $89.15\%$ for VR, and $88.72\%$ for SC, while the corresponding 3D mAP is $77.57\%$ for PP, $78.67\%$ for PR, $84.37\%$ for PVR, $85.66\%$ for VR, and $81.82\%$ for SC.

\paragraph{Attacks.}
We use PhyAdv~\cite{tu2020physically} as our baseline white-box attack with its original parameters: a maximum global offset of $0.1m$ in the $x$ and $y$ directions, no offset in the $z$ direction ($\vc = (0.1m, 0.1m, 0.0m)$), a Laplacian loss weight $\lambda = 0.001$, and the Adam optimizer with a learning rate of $\gamma = 0.005$. The 3D adversarial mesh is initialized as a unit sphere with 162 vertices and 320 triangular faces, scaled by $\vb = (0.7m, 0.7m, 0.5m)$.
By default, our attack uses gradient descent with misdetection loss (\ref{eq:full-loss-logit}), a sphere level $\nu = 2$, and a maximum global offset $\vc = (0.1m, 0.1m, 0.0m)$. The sphere level $\nu = 0, 1, 2, 3$ corresponds to vertex counts $||V|| = 12, 42, 162, 642$. Step sizes are: $\epsilon = 0.005$ for PP, $\epsilon = 0.05$ for PR, and $\epsilon = 0.0005$ for PVR, VR, and SC.

\paragraph{Rendering.}
For LiDAR simulation, we use the KITTI setup with a Velodyne HDL-64E sensor and a vertical resolution of about 0.4 degrees, covering elevation angles from +2° to -24.8°. To focus on adversarial meshes above car rooftops, we keep only the top 10 lines, from +2° to approximately -2.19°.
We use the algorithm from \cite{engelmann2017samp} for vehicle reconstruction. Since the code is not available, we train a TSDF using car models from the ApolloScape 3D dataset~\cite{song2019apollocar3d}. We then designate the top $0.2$ meters of the reconstructed surface as the car’s rooftop and place the adversarial patch at its center.

\section{Ablation for Key factors}

We examine the impact of key factors, \ie, the optimizer, initial mesh, and misdetection loss. For the ablation of initial mesh and misdetection loss, we selected PP, PR, and VR. This choice allows us to compare the differences between one-stage and two-stage networks, as well as point-based and voxel-based networks.

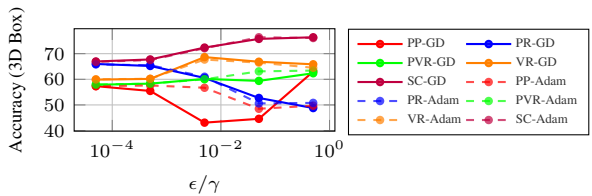
\begin{figure}[htp]
\centering
\begin{tabular}{c}
\begin{tikzpicture}
\begin{axis}[height=.35\columnwidth,
width=.6\columnwidth,
	xlabel=$\epsilon/ \gamma$ ,ylabel=Accuracy (3D Box),  
        xmode=log,
        legend columns=2,
	legend style={at={(1.05,1.00)},anchor=north west,nodes={scale=0.7, transform shape}}]	
        \addplot[red,mark=*]
        coordinates{
        (0.5,63.01) (0.05,44.58) (0.005,43.09) (0.0005,55.46) (0.00005,57.33) 

        };\leg{PP-GD}
        \addplot[blue,mark=*]
        coordinates{
        (0.5,48.78) (0.05,52.72) (0.005,60.54) (0.0005,65.20) (0.00005,65.98) 

        };\leg{PR-GD}
        \addplot[green,mark=*]
        coordinates{
        (0.5,62.36) (0.05,59.38) (0.005,60.14) (0.0005,58.31) (0.00005,57.86) 
        };\leg{PVR-GD}
        \addplot[orange,mark=*]
        coordinates{
        (0.5,65.85) (0.05,66.85) (0.005,68.65) (0.0005,60.16) (0.00005,59.87) 
        };\leg{VR-GD}
        \addplot[purple,mark=*]
        coordinates{
        (0.5,76.39) (0.05,75.76) (0.005,72.37) (0.0005,67.74) (0.00005,66.94) 
        };\leg{SC-GD}

        \addplot[red,dash,mark=*]
        coordinates{
        (0.5,49.7)(0.05,48.5)(0.005,56.7)(0.0005,57.6)(0.00005,57.3)
        };\leg{PP-Adam}
        \addplot[blue,dash,mark=*]
        coordinates{
        (0.5,50.8)(0.05,50.6)(0.005,61.0)(0.0005,65.5)(0.00005,65.8)
        };\leg{PR-Adam}
        \addplot[green,dash,mark=*]
        coordinates{
        (0.5,63.4)(0.05,63.1)(0.005,60.0)(0.0005,58.3)(0.00005,58.0)
        };\leg{PVR-Adam}
        \addplot[orange,dash,mark=*]
        coordinates{
        (0.5,64.5)(0.05,66.5)(0.005,67.7)(0.0005,60.2)(0.00005,60.0)
        };\leg{VR-Adam}
        \addplot[purple,dash,mark=*]
        coordinates{
        (0.5,76.1)(0.05,76.4)(0.005,72.1)(0.0005,67.5)(0.00005,66.8)
        };\leg{SC-Adam}
\end{axis}
\end{tikzpicture}
\end{tabular}

\caption{Ablation on step size $\epsilon$ for gradient descend or learning rate $\gamma$ for Adam.}
\label{fig:para-white}
\end{figure}

\paragraph{Influence of the Optimizer.}
To compare gradient descent and Adam optimizers in the white-box setting, we varied the step size $\epsilon$ for gradient descent and the learning rate $\gamma$ for Adam across five 3D object detection models. For two-stage models, the loss was constructed in the second stage. The results, shown in Figure \ref{fig:para-white}, indicate that Gradient Descent (GD) performs as well as or better than Adam. Consequently, GD is used for the remaining experiments. We also find that PR benefits from a larger step size, SC and VR perform better with a smaller step size, while PP has an optimal step size, and PVR shows little sensitivity to step size variations. In experiments, we use the optimal parameters for each model.


\paragraph{Influence of the Initial Mesh.}

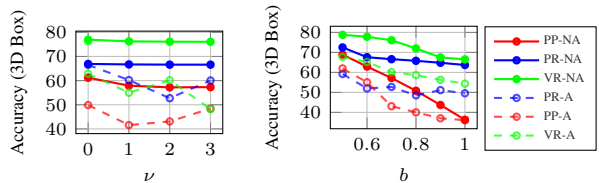
\begin{figure}[htp]
\centering
\begin{tabular}{c c}
    
        

\begin{tikzpicture}
\begin{axis}[height=.35\columnwidth,width=.42\columnwidth,
	xlabel=$\nu$,ylabel=Accuracy (3D Box),  
        legend columns=1,
	legend pos=outer north east]
	\addplot[red,mark=*]
        coordinates{
        (0,61.2)(1,57.9)(2,57.2)(3,57.3)
        };\leg{PP-NA}
        \addplot[blue,mark=*]
        coordinates{
        (0,66.9)(1,66.7)(2,66.6)(3,66.6)
        };\leg{PR-NA}
        \addplot[green,mark=*]
         coordinates{
         (0, 76.94)(1, 76.25)(2, 76.10)(3, 76.06)
         };\leg{VR-NA}

        \addplot[red,dash]
        coordinates{
        (0, 49.86) (1, 41.59) (2,43.09) (3, 48.41) 

        };\leg{PP-A}
        \addplot[blue,dash]
        coordinates{
        
        (0, 66.51) (1, 60.21) (2,52.72) (3, 60.05) 
        };\leg{PR-A}
        \addplot[green,dash]
         coordinates{
         (0.00005,76.31) 
    
         (0, 62.88) (1, 54.98) (2,60.16)  (3, 48.20) 
         };\leg{VR-A}
         \legend{}
         
\end{axis}
\end{tikzpicture} & 

        
\begin{tikzpicture}
\begin{axis}[height=.35\columnwidth,width=.42\columnwidth,
	xlabel=$b$,ylabel=Accuracy (3D Box),  
        legend columns=1,
	legend style={at={(1.05,1.0)},anchor=north west,nodes={scale=0.7, transform shape}}]
	\addplot[red,mark=*]
        coordinates{
        (0.5,68.9)(0.6,62.9)(0.7,57.2)(0.8,50.8)(0.9,43.7)(1.0,36.3)
        };\leg{PP-NA}
        \addplot[blue,mark=*]
        coordinates{
        (0.5,72.5)(0.6,67.6)(0.7,66.6)(0.8,65.8)(0.9,64.8)(1.0,63.7)
        };\leg{PR-NA}
         \addplot[green,mark=*]
         coordinates{
         (0.5,78.8)(0.6,77.8)(0.7,76.1)(0.8,72.0)(0.9,67.5)(1.0,66.5)
         };\leg{VR-NA}

         \addplot[blue,dash]
        coordinates{
        (0.5, 59.21) (0.6, 51.97) (0.7,52.72) (0.8, 48.52) (0.9, 51.12) (1.0, 49.55) 
        };\leg{PR-A}
        \addplot[red,dash]
        coordinates{
        (0.5, 61.98) (0.6, 55.05) (0.7,43.09) (0.8, 40.01) (0.9, 37.05) (1.0, 36.09) 

        };\leg{PP-A}
        \addplot[green,dash]
         coordinates{
         (0.5, 67.75) (0.6, 65.24) (0.7,60.16) (0.8, 58.63) (0.9, 56.29) (1.0, 54.35) 

         };\leg{VR-A}
\end{axis}
\end{tikzpicture} 
\end{tabular}
\caption{Ablation on sphere level $\nu$ and scale $\vb$. In the left figure, $\vb =(0.7m,0.7m,0.5m)$; in the right figure, $\vb = (bm, bm, 0.5m)$ with $\nu=3$.}
\label{fig:sphere-level}
\end{figure}

In this section, we assess the impact of the initial mesh on attack performance by varying two parameters: the sphere level $\nu$, which affects the number of vertices, and the scale $\vb$, which sets the mesh radius. We evaluate two scenarios: placing the sphere mesh on top of the car to test its effect on detection and optimizing the mesh to create an adversarial one. For two-stage models, we use (\ref{eq:full-loss-logit}) as the misdetection loss in $\cL$. As shown in Figure \ref{fig:sphere-level}, PP is the most sensitive to mesh scale, while PR is the least. 
The impact of the sphere level varies depending on the model during the attack. Our findings indicate that a larger initial mesh improves attack performance, but for physical reliability, we set $b=0.7$. The optimal choice for $\nu$ is $\nu=2$.

\paragraph{Influence of the Misdetection Loss.}

We examine various loss functions using GD with $\epsilon = 0.05$ on PR in a white-box setting, focusing on changes to $\cL$ in the loss function. As shown in Table \ref{tab:loss-frozen-single}, BEV mAP is more difficult to attack compared to 3D mAP, with an ASR gap of over $\sim 15\%$ for PP and VR, and about $\sim 10\%$ for PR. VR remains stable despite variations in misdetection loss. Generally, MR loss performs better than ML loss. There is no clear optimal choice, but MR losses from (\ref{eq:full-loss}), (\ref{eq:full-loss1}) and (\ref{eq:full-loss-logit}) are strong options. Misdetection loss is usually more effective when constructed in Stage 1, though exceptions exist.

\begin{table}[h]
\centering
\scriptsize
\setlength{\tabcolsep}{1.5pt}
\begin{tabular}{lr|cc|cc|cc|cc|cc}
\toprule
\multicolumn{2}{c|}{\mr{3}{\rotatebox{90}{~Loss}}}&\multicolumn{2}{c|}{\mr{2}{PP~\shortcite{lang2019pointpillars}}}&\multicolumn{4}{c|}{PR~\shortcite{shi2019pointrcnn}}&\multicolumn{4}{c}{VR~\shortcite{deng2021voxel}}\\\cmidrule{5-12}
& && &\multicolumn{2}{c|}{Stage 1 (S1)}&\multicolumn{2}{c|}{Stage 2 (S2)}&\multicolumn{2}{c|}{Stage 1 (S1)}&\multicolumn{2}{c}{Stage 2 (S2)}\\ \cmidrule{3-12}
& &BEV &3D Box&BEV &3D Box&BEV &3D Box &BEV &3D Box&BEV &3D Box\\\midrule

\mr{3}{\rotatebox{90}{~ML}}
&(\ref{eq:iou-loss})&	15.60 &	42.92 &	23.02 &	29.56 &	15.01 &	23.59 &	15.07 &	29.84 &	15.13 &	30.52 \\
&(\ref{eq:full-loss}) &	16.26 &		44.32 &	21.56 &		30.07 &	21.16 &	29.83 &	15.19 &	\textbf{30.17} &	14.93 &	\textbf{30.63} \\
&(\ref{eq:full-loss1})&	16.38 &	43.81 &	24.44 &	32.32 &	22.75 &	29.75 &	15.13 &	29.95 &	\underline{15.16} &	\underline{30.53} 
\\\midrule
\mr{4}{\rotatebox{90}{~MR}}
&(\ref{eq:logit-loss}) &\underline{24.96} &	\underline{45.43} &	24.29 &	34.61 &	24.45 &	34.00 &	15.14 &	30.09 &	15.10 &	29.86 \\
&(\ref{eq:full-loss})&	\textbf{25.50} &	\textbf{46.50} &	24.37 &	34.20 &	24.43 &	34.15 &	\textbf{15.22} &	\textbf{30.17} &	15.03 &	29.88 

\\

&(\ref{eq:full-loss1})&	23.25 &	43.95 &	\textbf{27.14} &	\textbf{38.36} &	23.89 &	\underline{35.22} &	{15.19} &	30.07 &	15.03 &	29.79 \\
&(\ref{eq:full-loss-logit})&	23.58 &	44.26 &	23.80 &	32.21 &	\textbf{26.74} &	\textbf{37.30} &	15.15 &	\underline{30.16} &	15.04 &	29.83 
\\

\midrule

\mr{2}{\rotatebox{90}{~C}}&(\ref{eq:full-loss}) &	17.68 &	43.69 &	25.27 &	36.34 &	\underline{25.56} &	34.94 &	\underline{15.20} &	30.13 &	15.05 &	29.88 \\
&(\ref{eq:full-loss1})&	16.91 &	43.10 &	\underline{25.73} &	\underline{37.62} &	25.00 &	34.77 &	15.16 &	29.95 &	\textbf{15.19} &	30.32   \\
\bottomrule
\end{tabular}
\caption{Comparison of misdetection loss designs. Bold indicates the best, and underlined denotes the second best.}
\label{tab:loss-frozen-single}
\end{table}

\section{Experimental Comparison}

\paragraph{White-Box Setting.}

In this section, we demonstrate the performance of adversarial meshes generated by our framework in white-box settings and their transferability across various models and datasets. We focus on two representative attacks using MR (\ref{eq:full-loss1}) and (\ref{eq:full-loss-logit}) since adversarial loss with logit shows better transferability in images~\cite{zhao2021success}. We investigate if this holds for 3D objects.

\begin{table}[htbp]
    \centering
    \scriptsize
\setlength{\tabcolsep}{1pt}
    \begin{tabular}{l|l |cc|cc|cc|cc|cc}
    \toprule
\mr{2}{\rotatebox{90}{Dataset}}    &    \mr{2}{Method}& 
\multicolumn{2}{c|}{PP~\shortcite{lang2019pointpillars}}&
\multicolumn{2}{c|}{PR~\shortcite{shi2019pointrcnn}}&
\multicolumn{2}{c|}{PVR~\shortcite{shi2020pv}}&
\multicolumn{2}{c|}{VR~\shortcite{deng2021voxel}}&
\multicolumn{2}{c}{SC~\shortcite{yan2018second} }  \\\cmidrule{3-12}

& &BEV &3D Box&BEV &3D Box&BEV &3D Box&BEV &3D Box&BEV &3D Box\\\midrule
\mr{4}{\rotatebox{90}{KITTI}}
&Vanilla&7.09 &26.29 &10.16 &15.34 &6.53 &21.50 &1.18 &11.16 &1.31 &9.62 \\
&PhyAdv&	12.20 &	36.25 &	15.50 &	22.35 &	2.34 &	6.99 &	13.11 &	21.47 &	3.34 &	12.66 \\
&MR(\ref{eq:full-loss1})
&	23.25 &	43.95 &	23.89 &	35.22 &	\textbf{13.99} &	\textbf{31.98} &	15.03 &	29.79 &	\textbf{7.77} &	\textbf{17.45}  \\
&MR(\ref{eq:full-loss-logit})
&	\textbf{23.58} &	\textbf{44.26} &	\textbf{26.74} &	\textbf{37.30} &	8.60 &	30.69 &	\textbf{15.04} &	\textbf{29.83} &	\textbf{7.77} &	17.15 
\\

\midrule
\mr{4}{\rotatebox{90}{CarLA}}
&Vanilla& 	53.55& 	57.37  & 	33.52& 	48.28  & 	\textbf{37.92}& 	\textbf{41.21}  & 	\textbf{29.06}& 	32.01  & 	\textbf{43.87}& 	\textbf{46.98}  \\
&PhyAdv& 	60.23& 	66.58  & 	35.81& 	50.83  & 	25.38& 	27.88  & 	27.25& 	29.57  & 	38.82& 	41.25  \\
&MR(\ref{eq:full-loss1}) & 	67.85& 	72.35  & 	41.81& 	56.31  & 	35.54& 	38.66  & 	29.02& 	\textbf{32.02}  & 	43.00& 	46.09  \\
&MR(\ref{eq:full-loss-logit})  & 	\textbf{68.50}& 	\textbf{72.80}  & 	\textbf{46.02}& 	\textbf{61.90}  & 	{37.56}& 	40.54  & 	28.95& 	31.86  & 	42.94& 	45.97  \\

\bottomrule
    \end{tabular}
    \caption{
    ASR of adversarial meshes for different models in a white-box setting on the KITTI dataset, with transferability tested on CarLA. Bold indicates the best performance.}
    \label{tab:white-box-trans-dataset}
\end{table}

\tikzset{mark options={mark size=1}}

\newcommand{\lossAblationScatterPlot}[8]{
\begin{tikzpicture}
\begin{axis}[height=.30\columnwidth,width=.5\columnwidth,
	xlabel=#1,ylabel=#2, 
        ymin=#3, ymax=#4,
        xmin=#7,xmax=#8,
        #6,
        scatter/classes={
            pointpillar={color={rgb,255:red,31;green,119;blue,180}},  
            pointrcnn={color={rgb,255:red,255;green,127;blue,14}},  
            pointrcnn_s1={mark=square*, color={rgb,255:red,44;green,160;blue,44}}, 
            voxel_rcnn_car={color={rgb,255:red,214;green,39;blue,40}},
            voxel_rcnn_car_s1={mark=square*, color={rgb,255:red,148;green,103;blue,189}},
            pvrcnn={color={rgb,255:red,140;green,86;blue,75}},
            second={color={rgb,255:red,227;green,119;blue,194}}
        }]
]
    \addplot[
        scatter,only marks,
        scatter src=explicit symbolic
    ]
    table[meta=label] {#5};
    
\end{axis}
\end{tikzpicture}
}

\newcommand{\transferabilityScatterPlot}[7]{
\begin{tikzpicture}
\begin{axis}[height=.30\columnwidth,width=.5\columnwidth,
	xlabel=#1,  
        ymin=#2, ymax=#3,
        xmin=#6,xmax=#7,
        #5,
        scatter/classes={
            pointpillar={color={rgb,255:red,31;green,119;blue,180}},  
            pointrcnn={color={rgb,255:red,255;green,127;blue,14}},      
            pointrcnn_s1={mark=square*, color={rgb,255:red,44;green,160;blue,44}}, 
            voxel_rcnn_car={color={rgb,255:red,214;green,39;blue,40}},
            voxel_rcnn_car_s1={mark=square*, color={rgb,255:red,148;green,103;blue,189}},
            pvrcnn={color={rgb,255:red,140;green,86;blue,75}},
            second={color={rgb,255:red,227;green,119;blue,194}}
        }]
]
    \addplot[
        scatter,only marks,
        scatter src=explicit symbolic
    ]
    table[meta=label] {#4};
\end{axis}
\end{tikzpicture}
}

\newcommand{\lossAblationScatterPlotLast}[8]{
\begin{tikzpicture}
\begin{axis}[height=.30\columnwidth,width=.5\columnwidth,
	xlabel=#1,ylabel=#2, 
        ymin=#3, ymax=#4,
        xmin=#7,xmax=#8,
        #6,
        scatter/classes={
            pointpillar={color={rgb,255:red,31;green,119;blue,180}},  
            pointrcnn={color={rgb,255:red,255;green,127;blue,14}},  
            pointrcnn_s1={mark=square*, color={rgb,255:red,44;green,160;blue,44}}, 
            voxel_rcnn_car={color={rgb,255:red,214;green,39;blue,40}},
            voxel_rcnn_car_s1={mark=square*, color={rgb,255:red,148;green,103;blue,189}},
            pvrcnn={color={rgb,255:red,140;green,86;blue,75}},
            second={color={rgb,255:red,227;green,119;blue,194}}
        }]
]
    \addplot[
        scatter,only marks,
        scatter src=explicit symbolic
    ]
    table[meta=label] {#5};
\end{axis}
\end{tikzpicture}
}

\newcommand{\transferabilityScatterPlotLast}[7]{
\begin{tikzpicture}
\begin{axis}[height=.30\columnwidth,width=.5\columnwidth,
	xlabel=#1,  
        ymin=#2, ymax=#3,
        xmin=#6,xmax=#7,
        #5,
        scatter/classes={
            pointpillar={color={rgb,255:red,31;green,119;blue,180}},  
            pointrcnn={color={rgb,255:red,255;green,127;blue,14}},      
            pointrcnn_s1={mark=square*, color={rgb,255:red,44;green,160;blue,44}}, 
            voxel_rcnn_car={color={rgb,255:red,214;green,39;blue,40}},
            voxel_rcnn_car_s1={mark=square*, color={rgb,255:red,148;green,103;blue,189}},
            pvrcnn={color={rgb,255:red,140;green,86;blue,75}},
            second={color={rgb,255:red,227;green,119;blue,194}}
        }]
]
    \addplot[
        scatter,only marks,
        scatter src=explicit symbolic
    ]
    table[meta=label] {#4};
\end{axis}
\end{tikzpicture}
}

\begin{figure}[htp]
\centering
\begin{tabular}{c c}

\lossAblationScatterPlot{ASR (3D Box)}{$L_2$ Norm}{0}{6.5}{fig/loss_ablation_scatter/l2_norm_vs_3d_asr.csv}{}{25}{50} 
& \transferabilityScatterPlot{ASR (3D Box)}{0}{6.5}{fig/transferability_scatter/l2_norm_vs_3d_asr.csv}{}{5}{60}    \\ 
\lossAblationScatterPlot{ASR (3D Box)}{$\Phi$}{0}{0.2}{fig/loss_ablation_scatter/laplacian_vs_3d_asr.csv}{scaled y ticks = base 10:2}{25}{50}
& \transferabilityScatterPlot{ASR (3D Box)}{0}{0.2}{fig/transferability_scatter/laplacian_vs_3d_asr.csv}{scaled y ticks = base 10:2}{5}{60}   \\ 
\lossAblationScatterPlot{ASR (3D Box)}{Volume}{0}{1.5}{fig/loss_ablation_scatter/volume_vs_3d_asr.csv}{}{25}{50} 
& \transferabilityScatterPlot{ASR (3D Box)}{0}{1.5}{fig/transferability_scatter/volume_vs_3d_asr.csv}{}{5}{60}   \\ 

\lossAblationScatterPlotLast{ASR (BEV)}{$\text{Area}$}{1.2}{2.0}{fig/loss_ablation_scatter/projected_volume_vs_bev_asr.csv}{}{10}{30}
& \transferabilityScatterPlotLast{ASR (BEV)}{1.2}{2.0}{fig/transferability_scatter/projected_volume_vs_bev_asr.csv}{}{5}{25}   \\ 
\multicolumn{2}{c}{
\begin{tikzpicture}
\begin{axis}[
    hide axis,
    xmin=0, xmax=1,
    ymin=0, ymax=1,
    legend columns=7,
    legend style={draw=none, column sep=1ex, at={(0.5,-0.3)}, anchor=north},
    legend entries={PP, PR-S2, PR-S1, VR-S2, VR-S1, PV, SC}
]
    \addlegendimage{color={rgb,255:red,31;green,119;blue,180}, only marks}
    \addlegendimage{color={rgb,255:red,255;green,127;blue,14}, only marks}
    \addlegendimage{mark=square*, color={rgb,255:red,44;green,160;blue,44}, only marks}
    \addlegendimage{color={rgb,255:red,214;green,39;blue,40},only marks}
    \addlegendimage{mark=square*, color={rgb,255:red,148;green,103;blue,189}, only marks}
    \addlegendimage{color={rgb,255:red,140;green,86;blue,75}, only marks}
    \addlegendimage{color={rgb,255:red,227;green,119;blue,194}, only marks}
\end{axis}
\end{tikzpicture}} \\
\end{tabular}
\caption{
3D and BEV box ASR plotted with metrics: Volume, $L_2$ norm, Laplacian smoothness ($\Phi$), and projected area. Rows share y-axes. Left: white-box ASR (Table~\ref{tab:loss-frozen-single}); Right: transferability ASR (Table~\ref{tab:white-box-transferability}). Colors indicate target models (used for both generation and testing in white-box; testing only in transferability).}
\label{fig:distor-asr}
\end{figure}
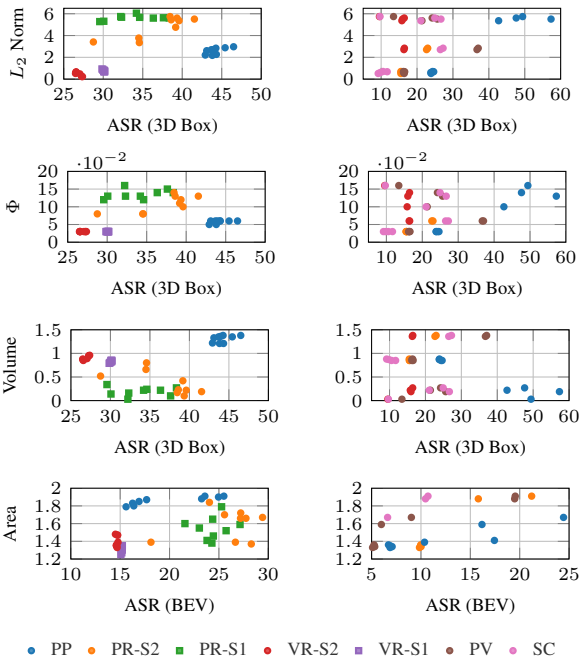


The results for the KITTI dataset in Table \ref{tab:white-box-trans-dataset} demonstrate that our attack significantly outperforms PhyAdv, especially on PP, PR, and PVR. Notably, on PVR, PhyAdv performs worse than Vanilla, which uses a spherical mesh placed on cars.
For the CarLA dataset, Table \ref{tab:white-box-trans-dataset} shows that our adversarial mesh transfers well across datasets, with models being highly affected by objects on top of cars. Vanilla achieves the highest ASR on PVR and SC, while our attack outperforms PhyAdv, particularly against PP and PR.

\paragraph{Transferability.} According to Table \ref{tab:white-box-transferability}, adversarial examples generated using PP show the highest transferability. Generally, BEV mAP is harder to compromise than 3D mAP, with VR being the most robust and PP the most vulnerable. Unlike image classification, using logits does not improve transferability for LiDAR-based detectors. Using MR  (\ref{eq:full-loss1}) as the misdetection loss improves transferability when attacking the second stage of PR, but not for VR. Adversarial examples generated with VR show the lowest transferability, indicating that models easier to attack produce more transferable adversarial examples.

To explore how invisibility affects white-box attacks and transferability, we measure four invisibility metrics. In Figure \ref{fig:distor-asr}, the first column of, derived from Table \ref{tab:loss-frozen-single}, shows no clear relationship between these metrics and ASR. However, the second column, from Table \ref{tab:white-box-transferability}, reveals that while $L_2$ norm, Laplacian smoothness $\phi$ and 3D box volume have an inconsistent impact on transferable 3D ASR, a larger BEV area consistently improves transferable BEV ASR. Therefore, to enhance transferability, increasing the BEV area of the adversarial mesh during optimization is beneficial.

\begin{table}[]
\scriptsize
\setlength{\tabcolsep}{0.3pt}
    \centering
    \begin{tabular}{l|l|l|cc|cc|cc|cc|cc}
    \toprule
Source&\mr{2}{\rotatebox{90}{Stage}}  &\mr{2}{Loss} &
\multicolumn{2}{c|}{PP~\shortcite{lang2019pointpillars}}&
\multicolumn{2}{c|}{PR~\shortcite{shi2019pointrcnn}}&
\multicolumn{2}{c|}{PVR~\shortcite{shi2020pv}}&
\multicolumn{2}{c|}{VR~\shortcite{deng2021voxel}}&
\multicolumn{2}{c}{SC~\shortcite{yan2018second} }  \\\cmidrule{4-13}
Model &&&BEV &3D Box&BEV &3D Box&BEV &3D Box&BEV &3D Box&BEV &3D Box\\\midrule
\mr{2}{{PP}}
  & - & MR(\ref{eq:full-loss1})
  & 	-& 	-  & 	\textbf{21.21}& 	\textbf{22.97}  & 	\textbf{19.55}& 	\textbf{37.04}  & 	\textbf{3.30}& 	\textbf{16.48}  & 	\textbf{10.71}& 	\textbf{27.25}  \\
  & - & MR(\ref{eq:full-loss-logit})
    & 	-& 	-  & 	\underline{15.81}& 	\underline{22.65}  & 	\underline{19.49}& 	\underline{36.70}  & 	\underline{3.19}& 	16.34  & 	\underline{10.46}& 	26.49  \\

  \midrule
  
\mr{4}{{PR}}  
    &\mr{2}{\rotatebox{90}{S1}}

   &MR(\ref{eq:full-loss1})
      & 	\underline{17.44}& 	\underline{49.40}  & 	-& 	-  & 	3.14& 	13.45  & 	0.89& 	9.53  & 	1.66& 	9.68  

      \\
     & &MR(\ref{eq:full-loss-logit})

& 	16.18& 	47.62  & 	-& 	-  & 	5.99& 	24.21  & 	1.52& 	\underline{16.44}  & 	3.92& 	24.94  
      \\

      \cmidrule{2-13}
   &\mr{2}{\rotatebox{90}{S2}}

   &MR(\ref{eq:full-loss1})
      & 	\textbf{24.44}& 	\textbf{57.32}  & 	-& 	-  & 	9.03& 	25.63  & 	1.75& 	15.97  & 	6.62& 	\underline{26.66}  \\

 & &MR(\ref{eq:full-loss-logit})

      & 	10.36& 	42.72  & 	-& 	-  & 	4.23& 	21.37  & 	0.84& 	15.78  & 	1.93& 	21.06

      \\

      \midrule

\mr{4}{{VR}}  
    &\mr{2}{\rotatebox{90}{S1}}
 &MR(\ref{eq:full-loss1})
      & 	7.08& 	24.64  & 	10.01& 	15.60  & 	5.32& 	16.56  & 	-& 	-  & 	1.79& 	11.69  

      \\
 &&MR(\ref{eq:full-loss-logit})
      & 	6.87& 	24.34  & 	9.85& 	15.51  & 	5.10& 	16.23  & 	-& 	-  & 	1.68& 	10.55  

      \\

      \cmidrule{2-13}
   &\mr{2}{\rotatebox{90}{S2}}

   &MR(\ref{eq:full-loss1})
      & 	6.88& 	23.84  & 	9.96& 	15.72  & 	5.26& 	16.24  & 	-& 	-  & 	1.65& 	9.79  
      \\
  &&MR(\ref{eq:full-loss-logit})
      & 	6.71& 	23.90  & 	9.98& 	15.49  & 	5.25& 	16.43  & 	-& 	-  & 	1.65& 	9.18  

      \\

\bottomrule
    \end{tabular}
    \caption{Transferability of adversarial meshes: generated against source models (columns) and tested on target models (rows). Bold indicates the best, and underlined denotes the second best.}
    \label{tab:white-box-transferability}
\end{table}




\begin{table}[htpb]
    \centering
    \scriptsize
\setlength{\tabcolsep}{1pt}
    \begin{tabular}{l|l |cc|cc|cc|cc|cc}
    \toprule
 Surrogate     & \mr{2}{\rotatebox{90}{Stage}} & 
\multicolumn{2}{c|}{PP~\shortcite{lang2019pointpillars}}&
\multicolumn{2}{c|}{PR~\shortcite{shi2019pointrcnn}}&
\multicolumn{2}{c|}{PVR~\shortcite{shi2020pv}}&
\multicolumn{2}{c|}{VR~\shortcite{deng2021voxel}}&
\multicolumn{2}{c}{SC~\shortcite{yan2018second}}  \\\cmidrule{3-12}
Model& &BEV &3D Box&BEV &3D Box&BEV &3D Box&BEV &3D Box&BEV &3D Box\\\midrule

PP & - 

& 	-& 	-  & 	\textbf{10.66}& 	\underline{16.44}  & 	\underline{7.81}& 	\underline{22.58}  & 	1.32& 	11.09  & 	1.98& 	\underline{15.27}  \\

\midrule
\mr{2}{PR}&  S1 
& 	\underline{15.09}& 	\underline{41.77}  & 	-& 	-  & 	\underline{7.81}& 	\textbf{24.43}  & 	\textbf{2.02}& 	\textbf{15.58}  & 	\underline{4.28}& 	15.11 \\
&S2  
& 	\textbf{17.51}& 	\textbf{47.41}  & 	-& 	-  & 	\textbf{7.83}& 	21.78  & 	\underline{1.72}& 	\underline{14.79}  & 	\textbf{5.53}& 	\textbf{17.90}  \\
 \midrule

\mr{2}{PVR} &S1
& 	7.07& 	26.50  & 	10.14& 	16.34  & 	-& 	-  & 	1.18& 	11.24  & 	1.21& 	9.10  \\
&S2
& 	6.83& 	25.37  & 	\underline{10.22}& 	\textbf{16.51}  & 	-& 	-  & 	1.20& 	11.29  & 	1.23& 	9.38  \\
 \midrule

\mr{2}{VR}&S1 & 	7.13& 	26.45  & 	10.20& 	15.99  & 	6.48& 	21.34  & 	-& 	-  & 	1.44& 	10.27  
\\

&S2 & 	7.11& 	26.44  & 	10.16& 	16.18  & 	6.38& 	21.35  & 	-& 	-  & 	1.26& 	8.98  \\

\bottomrule
    \end{tabular}
    \caption{ASR of adversarial meshes in a black-box setting, with surrogate models in the columns and target models in the rows. Bold indicates the best, and underlined denotes the second best.}
    \label{tab:black-box}
\end{table}
\paragraph{Black-Box Setting.}

According to Table \ref{tab:black-box}, we present the results of using PP, PR, PVR, and VR as source models ($f$) and assessing their performance against these models, as well as SC, as target models ($g$) in a black-box setting. The results show that PP is the most effective source model. Conversely, PVR and VR as source models perform poorly against SC as the target model, while VR is the most difficult target model when PP and PR are used as sources.

\begin{figure}[t]
\centering

\includegraphics[trim={12mm 32mm 12mm 45mm},clip,width=.45\textwidth]{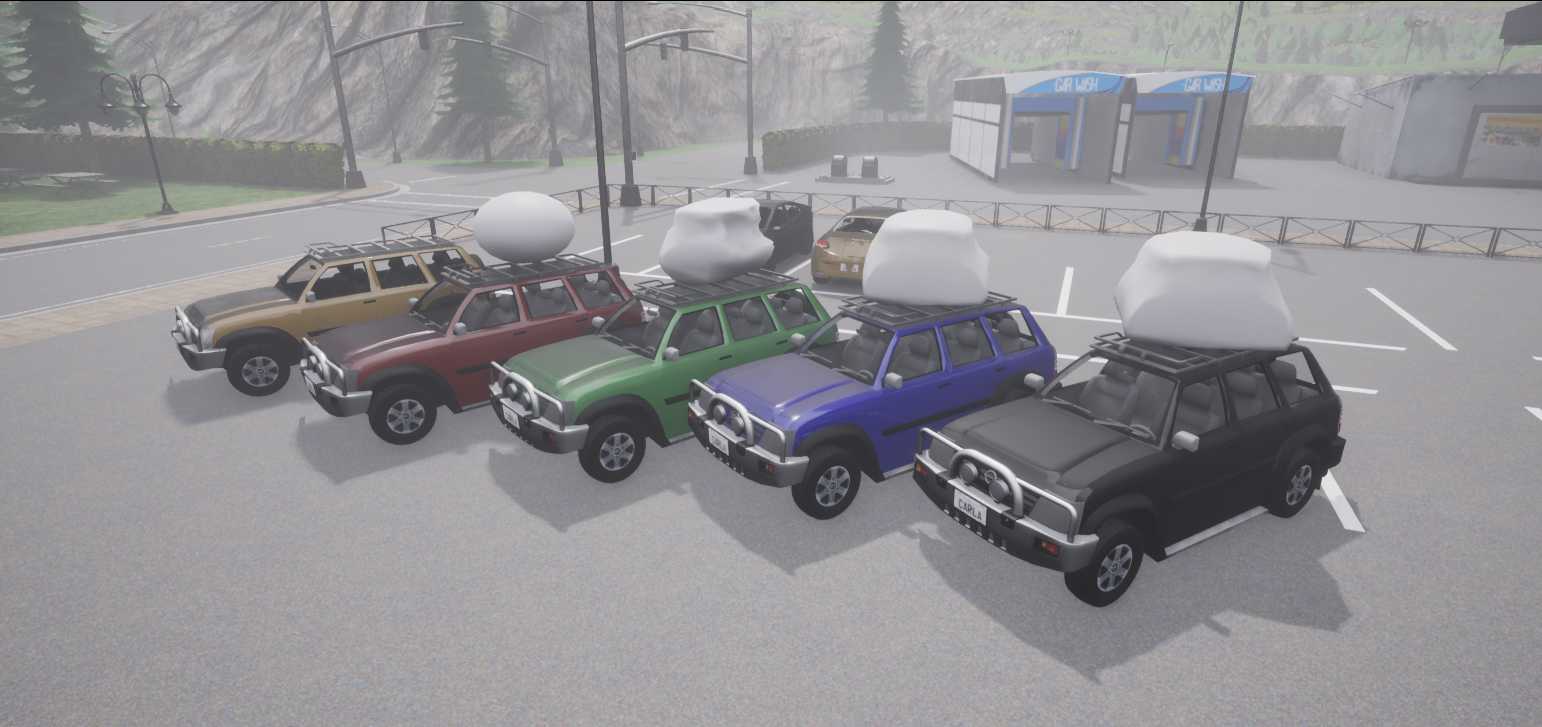}

\caption{As shown in the figure. From left to right, we construct an adversarial vehicle for clean, vanilla, PhyAdv, MR(\ref{eq:full-loss-logit}), and MR(\ref{eq:full-loss1}), respectively.}

\label{fig:carla-fig}
\end{figure}

\begin{table}[h!]
    \centering
    \scriptsize
\setlength{\tabcolsep}{5pt}
\begin{tabular}{l|cccc|cccc}
\toprule
 & \multicolumn{4}{c|}{Clean} & \multicolumn{4}{c}{Noisy} \\

 &CutIn&FLVO & FLV & OFLV &CutIn&FLVO & FLV & OFLV  \\
\midrule

Benign & 0.61 & 0.71 & 0.67 & 0.35 & 0.73 & 0.70 & 0.70 & 0.51 \\ 
Vanilla & 0.58 & 0.52 & 0.56 & 0.36 & 0.41 & 0.34 & 0.37 & 0.33 \\
PhyAdv  & 0.55 & 0.52 & 0.51 & 0.34  & 0.33 & 0.34 & 0.30 & 0.32 \\ 
MR(\ref{eq:full-loss1}) &\textbf{ 0.33} & 0.44 & 0.37 & \textbf{0.23}   & 0.19 & 0.28 & 0.21 & \textbf{0.21} \\
MR(\ref{eq:full-loss-logit}) & 0.36 & \textbf{0.42} & \textbf{0.36} & \textbf{0.23}
  & \textbf{0.17} & \textbf{0.26} & \textbf{0.20} & \textbf{0.21} \\
\bottomrule

\end{tabular} 

\caption{The mAP@0.7 results on CutIn, FLVWO, FLV, and OFLV scenarios using PointPillar, with confidence > 0.4. 'Clean' denotes CarLA-generated scenes without noise; 'Noisy' includes added simulation noise.}
\label{tab:carla-sim}
\end{table}

\paragraph{CarLA Simulation Scenarios.}

We generate adversarial objects from 3D point clouds in the KITTI dataset, convert them into mesh models, and deploy them in the CarLA simulator (see Figure \ref{fig:carla-fig}). Their impact is evaluated across predefined driving scenarios using the ScenarioRunner framework~\footnote{\url{https://github.com/carla-simulator/scenario_runner}}, focusing on four safety-critical cases: Cut-In, FLV, OFLV, and FLVWO. 
In CutIn, a vehicle suddenly merges into the ego vehicle’s lane, testing the system’s ability to perceive and react quickly. FLV (Follow Leading Vehicle) involves the ego vehicle trailing another under normal conditions, assessing persistent and reliable perception. In OFLV (Other Follow Leading Vehicle), the ego vehicle must overtake an obstructing car to continue following a farther target, challenging perception under occlusion. Finally, FLVWO (Follow Leading Vehicle With Obstacle) combines following behavior with unexpected obstacle avoidance, requiring both timely and persistent perception.
Each scenario is tested under two conditions, \ie Clean (no added noise) and Noisy (Gaussian noise with $\sigma = 0.2$), to simulate different sensor conditions. We compare five object types: no attack (benign), a basic sphere (vanilla), adversarial objects generated by phyADV, and two adversarial variants optimized for misdetection loss (MR(\ref{eq:full-loss1}) and MR(\ref{eq:full-loss-logit})). Quantitative results, as Table \ref{tab:carla-sim}, show that MR(\ref{eq:full-loss-logit}) outperforms in CarLA simulation scenarios. 

\section{Challenges}
Compared to digital adversarial attacks on LiDAR-based systems, physically realizable adversarial object attacks remain underexplored, particularly in terms of implementation challenges. Addressing this problem is crucial for safety-critical applications such as autonomous driving, where real-world robustness and compliance with safety regulations are essential.
Several open challenges persist. For example, commonly used constraints such as bounding box limits and surface smoothness (e.g., via Laplacian loss) do not adequately capture physical feasibility. There is a need for more effective constraints that better reflect 3D object plausibility while remaining differentiable for optimization. Additionally, while we explore various formulations for adversarial misdetection loss, practical implementation, especially for two-stage detectors, remains difficult. Backpropagating through refinement stages is non-trivial. Besides, current designs often rely only on coarse metrics like IoU, logits, or classification probabilities. Our framework includes a feasible implementation, but we believe improved loss formulations incorporating richer features could further enhance attack effectiveness.

Another underexplored issue is the reflectivity of adversarial objects. In real-world autonomous driving datasets such as KITTI and nuScenes, point clouds are typically stored in an $N \times 4$ format, where the first three channels represent spatial coordinates, and the fourth channel encodes reflectivity, \ie, the intensity of the laser return. We neutralize reflectivity in our evaluation by setting it to zero across all points, ensuring a pure geometric baseline and isolating the effects of spatial structure on model behavior. Nevertheless, while reflectivity can enhance classification performance, its value distribution varies significantly across different LiDAR sensors, leading to two key challenges: (1) models trained on different datasets may learn inconsistent reflectivity patterns, and (2) existing work's focus is on analyzing geometric structure, not sensor-specific intensity cues.

\section{Conclusion}

In this work, we propose a unified, device-agnostic framework for generating and evaluating physically realizable adversarial objects in LiDAR-based 3D object detection. Our framework supports diverse model architectures, attack settings, and loss designs, offering flexibility while enabling reproducible benchmarking. By abstracting the key components of physical attacks and validating their transferability from simulation to the real world, we demonstrate that simulation can serve as an effective proxy for physical feasibility. We also highlight underexplored factors, such as reflectivity and constraint formulation, that influence attack success. Alongside our open-source code and standardized evaluation protocol, we aim to accelerate progress and foster deeper understanding of adversarial robustness in real-world perception systems.





{
\footnotesize
\bibliography{main}
}
\end{document}